\documentclass[conference]{IEEEtran}
\IEEEoverridecommandlockouts
\usepackage{cite}
\usepackage{amsmath,amssymb,amsfonts}
\usepackage{algorithmic}

\usepackage{algorithm,algorithmic}
\usepackage{graphicx}
\usepackage{makecell}
\usepackage{textcomp}
\usepackage{tabularx}
\usepackage{MnSymbol,wasysym}
\usepackage[utf8]{inputenc}
\usepackage[main=english,vietnamese]{babel}
\usepackage{xcolor}
\def\BibTeX{{\rm B\kern-.05em{\sc i\kern-.025em b}\kern-.08em
    T\kern-.1667em\lower.7ex\hbox{E}\kern-.125emX}}
\begin{document}

\title{Non-Standard Vietnamese Word Detection and Normalization for Text--to--Speech
}

\author{
    \IEEEauthorblockN{
        \textbf{Huu-Tien Dang}\IEEEauthorrefmark{1}\IEEEauthorrefmark{2}, \textbf{Thi-Hai-Yen Vuong}\IEEEauthorrefmark{2}, \textbf{Xuan-Hieu Phan}\IEEEauthorrefmark{2}
    }
    \IEEEauthorblockA{\IEEEauthorrefmark{1}\textit{FPT Technology Research Institute, FPT University, Hanoi, Vietnam}}
    \IEEEauthorblockA{\IEEEauthorrefmark{2}\textit{VNU University of Engineering and Technology, Hanoi, Vietnam}}
    \IEEEauthorrefmark{1}tiendh7@fpt.com.vn \\\IEEEauthorrefmark{2}\{yenvth, hieupx\}@vnu.edu.vn
    
}


\maketitle

\begin{abstract}
Converting written texts into their spoken forms is an essential problem in any text--to--speech (TTS) systems. However, building an effective text normalization solution for a real--world TTS system face two main challenges: (1) the semantic ambiguity of non-standard words (NSWs), e.g., numbers, dates, ranges, scores, abbreviations, and (2) transforming NSWs into pronounceable syllables, such as URL, email address, hashtag, and contact name. In this paper, we propose a new two--phase normalization approach to deal with these challenges. First, a model--based tagger is designed to detect NSWs. Then, depending on NSW types, a rule--based normalizer expands those NSWs into their final verbal forms. We conducted three empirical experiments for NSW detection using Conditional Random Fields (CRFs), BiLSTM-CNN-CRF, and BERT-BiGRU-CRF models on a manually annotated dataset including 5819 sentences extracted from Vietnamese news articles. In the second phase, we propose a forward lexicon-based maximum matching algorithm to split down the hashtag, email, URL, and contact name. The experimental results of the tagging phase show that the average $F_1$ scores of the BiLSTM-CNN-CRF and CRF models are above 90.00\%, reaching the highest $F_1$ of 95.00\% with the BERT-BiGRU-CRF model. Overall, our approach has low sentence error rates, at 8.15\% with CRF and 7.11\% with BiLSTM-CNN-CRF taggers, and only 6.67\% with BERT-BiGRU-CRF tagger.
\end{abstract}

\begin{IEEEkeywords}
Text--to--speech normalization, non-standard word detection, Conditional Random Fields, BiLSTM-CNN-CRF, BERT-BiGRU-CRF.
\end{IEEEkeywords}

\section{Introduction} 

Text--to--speech (TTS) is the process of creating speech signals from texts. TTS can be used in various applications, such as news reading, call--bot, and film narration. In addition, voice assistants, e.g., Google Assistant, Apple Siri, Microsoft Cortana, and Amazon Alexa, all need a TTS module. In Vietnam, there are also several TTS systems for Vietnamese, such as Vbee, Zalo KiKi, and VinBigData ViVi.

\begin{table}[htbp]
\caption{Examples of converting written texts into spoken forms}
\begin{center}
\begin{tabularx}{0.49\textwidth}{|X|X|}
\hline
\textbf{Written form text}&\textbf{Spoken form text}\\
\hline

\begin{otherlanguage}{vietnamese}Ngày \textbf{31/3}, gần \textbf{92000} ca mắc mới \textbf{Covid-19} ở \textbf{Tp}. Hà Nội\quad\quad\quad  \quad(On March 31, nearly 92000 new cases of Covid-19 in Hanoi city)\end{otherlanguage} & \begin{otherlanguage}{vietnamese}Ngày \textbf{ba mươi mốt tháng ba} , gần \textbf{chín mươi hai nghìn} ca mắc mới \textbf{cô vít mười chín} ở \textbf{thành phố} Hà Nội\end{otherlanguage} \quad\quad\quad\quad\quad\quad\quad\quad\quad (On the thirty first of March , nearly ninety two thousand new cases
of Covid nineteen in Hanoi city)\\

\hline

\begin{otherlanguage}{vietnamese}Trong ngày \textbf{3/4}, có \textbf{3/4} xe được bán.\end{otherlanguage}\quad(On April 3, 3/4 cars sold out.) & \begin{otherlanguage}{vietnamese}Trong ngày \textbf{mùng ba tháng tư} , có \textbf{ba trên bốn} xe được bán .\end{otherlanguage} \quad \quad \quad(On April third , three out of four cars sold out .)\\

\hline
\begin{otherlanguage}{vietnamese}Ngày \textbf{3-1}, \textbf{ĐT VN 3-1 ĐT TQ}\end{otherlanguage} \quad (On January 3, Vietnam national football team 3-1 China national football team)& \begin{otherlanguage}{vietnamese}Ngày \textbf{mùng ba tháng một} , \textbf{đội tuyển Việt Nam ba một đội tuyển Trung Quốc}\end{otherlanguage}\quad \quad\quad \quad\quad \quad  \quad \quad(On January third , Vietnam national football team three one China national football team)\\
\hline
\end{tabularx}
\label{tableI}
\end{center}
\end{table}

Among the important components of TTS, text normalization is the first processing step that helps to convert written texts into their spoken forms. Table I shows several examples of inputs and outputs of TTS text normalization. The written word, e.g., ``92000'', must be converted into \begin{otherlanguage}{vietnamese}
``chín mươi hai nghìn''
\end{otherlanguage} (``ninety two thousand''), which is often called a \textit{non-standard word} (NSW)\cite{1}. In most writing genres, such as news articles and social media, NSWs may have different forms, including numbers, digit sequences, abbreviations, fractions, numeric range, email addresses, game scores, URLs, etc. Identifying these NSWs accurately is a challenging task. The appropriate verbalization of NSWs can be different depending on what it presents in the context. For example, ``2/3'' can be a fraction \begin{otherlanguage}{vietnamese}``hai phần ba''\end{otherlanguage} (``two third'') or a date \begin{otherlanguage}{vietnamese}``
mùng hai tháng ba''\end{otherlanguage} (``the second of March''). ``$911$'' can be a digit sequence ``\begin{otherlanguage}{vietnamese}chín một một\end{otherlanguage}'' (``nine one one'') or a number ``\begin{otherlanguage}{vietnamese}chín trăm mười một\end{otherlanguage}'' (``nine hundred eleven''). An abbreviation ``Hn'' can be expanded as ``\begin{otherlanguage}{vietnamese}Hôm nay\end{otherlanguage}'' (``Today'') or ``\begin{otherlanguage}{vietnamese}Hà Nội\end{otherlanguage}'' (``Hanoi''). NSWs tends to be more ambiguous than other standard words, and context is essential for identifying these NSWs.

Typically, the rule-based approaches identify NSWs using dictionary and regular expressions\cite{dinh2012,27}. A dictionary helps classify words in a sentence into standard (in-vocabulary) or non-standard words (out-of-vocabulary - OOV). Then, regular expressions are used to determine which class each word belongs to. The shortcoming of this approach is that we need to update the dictionary manually every time having new words, and there are many words in OOV that do not need to be expanded. Besides, using the regular expressions will not completely resolve the semantic ambiguities. For recent studies, neural-based models to identify NSW in the form of the sequence classification problem have achieved high accuracy\cite{NAACL2021, 16,Ro2022}. In Vietnam, Trang et al.\cite{trang2010,trang2019} use a Vietnamese pronounceable syllables dictionary to identify NSWs, and traditional machine learning methods include Decision Tree and Random Forest classifier to classify NSWs. The drawback of this approach is that it needs external resources and requires more effort to maintain and update.

In addition, breaking multi-syllable tokens in email addresses, URLs, contact names, or hashtags such as ``phongdaotao'', ``anhkhanh'' into individual syllable tokens ``phong dao tao'', ``anh khanh'' is an important task for analyzing and understanding user utterances for TTS systems. Humans with commonsense background knowledge find it easy but TTS systems will struggle to pronounce these complex tokens. In previous work, Zhang et al. \cite{11} proposed a semi-supervised method for URL segmentation with an RNN pre-trained on a knowledge graph. The model relies heavily upon massive labeled training data. Despite giving good performance, this method only focus on URLs and is slow in inference.

To overcome these two challenges, in this paper, we propose a new approach for Vietnamese TTS normalization, which consists of two main phases: (1) \textit{NSW detection} which identifies NSWs in sentences with a model-based tagger, and (2) \textit{NSW normalization} that uses a set of pre-defined specific rule classes to normalize these NSW tokens into their spoken forms. In order to break down URLs, email addresses, contact names, and hashtags, we propose a greedy-based algorithm called \textit{forward lexicon-based maximum matching} (FLMM). FLMM uses a sliding window and slides through the word. While sliding, FLMM looks up the window-based sub-string in a pre-built lexicon and matches the longest string. FLMM marks the boundary and split the word at the boundary. This algorithm is quite simple and easy to implement for both experiments and production.

Our work has three main contributions as follows. First, we discussed non-standard Vietnamese word detection task as a sequence tagging problem. Second, we propose a forward lexicon-based maximum matching algorithm to break down URLs, email addresses, contact names, and hashtags. Third, we benchmark three empirical models on a manually annotated dataset including 5819 sentences. The experiment results show that our method have achieved high accuracy and suggesting a promising approach to TTS normalization and production.

\section{Related Work}

Previous works on TTS normalization have been mainly hard-coded rules \cite{4,5,6}. TTS normalization in MITalk\cite{4} is one of the earliest text normalization modules for TTS. Bell Labs’ multilingual TTS system introduced the weighted finite-state transducer for text normalization\cite{5,6}. Additionally, machine learning models combined with hand-written grammar were proposed for specific NSW classes. Roark et al. \cite{8} reported the use of maximum entropy for classifying letter sequence and Sproat et al. \cite{9} proposed a pair character language model for abbreviation modeling.

Inspired by the recent successes of neural networks in various natural language processing tasks, many neural network models were proposed to solve the TTS normalization problem \cite{12,13,14,15,16,NAACL2021,Ro2022}. Sproat et al.\cite{12,13} proposed RNN-based architectures for text normalization along with an open-source corpus consisting of the corresponding pairs of written and spoken sentences. A follow-up study focused on the convolutional neural network models\cite{14}. Another approach is to treat TTS normalization as a machine translation problem\cite{15,26}. Tyagi et al.\cite{16} proposed a hybrid system that includes a tagging model with the tokenization mechanism that enables the system to learn the majority of the classes automatically combined with pre-coded linguistic knowledge classes for NSW expansion. Lai et al. \cite{NAACL2021} proposed a unified framework for building a single duplex system that can handle TTS normalization and inverse TTS normalization. More recently, Ro et al. \cite{Ro2022} present an experimental comparison of various Transformer-based models of TTS normalization.

In the context of Vietnamese TTS normalization, Trang et al.\cite{trang2010} first developed a Vietnamese NSWs taxonomy that includes $21$ types and proposed a Decision Tree classifier and letter language model for NSW classification and a hybrid normalizer for NSW expansion. In 2019, Trang et al.\cite{trang2019} proposed an approach with the Random Forest classifier to identify NSW and the hybrid method combining a sequence-to-sequence model and rule-based for abbreviations and loan word expansion. In 2012, a rule-based approach was presented in\cite{dinh2012} using regular expression and a decision list to categorize and expand NSW. In 2021, Cuc et al.\cite{cuc2021} proposed a multi-task end-to-end model based on a denoising auto-encoder to address the OOV problem and the verbalizing of abbreviations.

\section{The Proposed Approach}

\begin{table*}
\caption{\label{citation-guide}
A modified NSWs taxonomy for tagging model.
}
\centering
\resizebox{\textwidth}{!}{
\begin{tabular}{|l|l|l|l|}
 \hline
 \textbf{Group} & \textbf{Tags} & \textbf{Description} & \textbf{Examples}\\
 \hline
   & NTIM&  Time&1h20, 1:20, 1:20:30, 1h20p30s, 11', 1g20', 12h-13h\\\cline{2-4}
   & NDAT&Full date&13/12/2021, 12.12.2021, 12-12-2021, 
1-2/3/2021, 8/9-10/9/2021, 2/3/2021-2/3/2022 \\\cline{2-4}
   & NDAY&Day and month&17/02, 13-12, 13.12\\\cline{2-4}
   & NMON&Month and year&02/2021, 12-2021, 12/2021, 12.2021\\\cline{2-4}
   & NQUA&Quarter&Quý I/2020\\\cline{2-4}
   & NNUM&Number&12, 70.000, 70 000, 700.005,6, -100\\\cline{2-4}
 \textbf{\textit{Number}}  & NDIG&Telephone/Number as digit&0977-1293-12, (+84) 0966 6354 12, 065.743.659, 0974 763 278, 114\\\cline{2-4}
   & NSCR&Score&\begin{otherlanguage}{vietnamese}tỷ số 2-3, mùa giải 2018-2019\end{otherlanguage}\\\cline{2-4}
   & NRNG&Range&\begin{otherlanguage}{vietnamese}từ 2-3 ngày\end{otherlanguage}\\\cline{2-4}
   & NPER&Percent&20\%, 30\% 20-30\%\\\cline{2-4}
   & NFRC&Fraction&24/7, \begin{otherlanguage}{vietnamese}tỷ lệ 2/3\end{otherlanguage}\\\cline{2-4}
   &NVER&Version&\begin{otherlanguage}{vietnamese}CM 4.0, phiên bản Android 7.0\end{otherlanguage}\\\cline{2-4}
\hline
   & LABB&Abbreviation&UBND, HLV, ĐT\\\cline{2-4}
  \textbf{\textit{Letter}} &LWRD &Foreign word&Ronaldo, Messi, NATO\\\cline{2-4}
   & LSEQ&Read as sequence&TTS, ASR, WHO, VTV\\\cline{2-4}
 \hline 
    & URLE&Email, URL, hashtag, contact name&\#anhkhanh, \#hienho, phongdaotao@vnu.edu.vn\\\cline{2-4}
  \textbf{\textit{Other}} & MONEY &Currency&2\$, \$2, 1000VNĐ, 1000đ \\\cline{2-4}
   & ROMA&Roman numeral&I, II, III, V, VI, X\\\cline{2-4}
   & MEA&Measurement&100kg, 100g, 100 kg, 10km2, 30oC\\\cline{2-4}
\hline

\end{tabular}}
\label{tabII}
\end{table*}

This section describes our approach to non-standard Vietnamese word detection and normalization. Our two-phase approach consists of three main modules: (1) pre-processing, (2) non-standard word detection, and (3) non-standard word normalization. In the pre-processing module, we remove extra spaces, emojis, HTML tags, unspoken tokens and present the input sentence in vector spaces. Afterward, a sequence tagging model is trained to extract NSWs according to a defined list of labels in the second module. Each type of NSW should be expanded by different expanders. In the non-standard word normalization, the expanders are defined based on the grammatical structure characteristic of the each non-standard word type. We also propose a forward lexicon-based maximum matching algorithm to separate URL, hashtag, contact name, and email name into separate syllables for TTS. 

\subsection{Pre-processing}

The pre-processing process has three steps. First, the extra space, ASCII art, emojis (e.g.,  \smiley{}, \frownie{}), HTML entities (e.g., ``\&nbsp'', ``\&lt'', ``\&gt'', ``\&amp'', ``\&quot'') and unspoken words (e.g., ``:))'') are removed. Second, a regex-based tokenizer is then used to split punctuations such as dots, commas, ellipses, quotation marks, etc., into separate tokens. Third, the input sentence is embedded into dense vector spaces to obtain input representation.


\subsection{Non-Standard Word Detection}

The NSW detection problem is solved as a sequence tagging problem. Let $\mathcal{S}_{train} = \{(\textbf{p}_{i},\textbf{t}_{i})\}_{i=0}^{n}$ be a n-examples training set drawn from distribution $\mathcal{D_{\mathcal{X}\times \mathcal{L}}}$. $\mathcal{X}$ and $\mathcal{L}$ are representation vector spaces of sentences and sequence labels respectively, $\textbf{p}_i \in \mathcal{X}$ is representation vector of input pre-processed sentence $p_i$, $\textbf{t}_i \in \mathcal{L}$ is a sequence over the alphabet $L$ of labels. The task to use $\mathcal{S}_{train}$ to learn a function $f: \mathcal{X} \to \mathcal{L}$ to label the sequences in the test set $\mathcal{S}_{test} \subset \mathcal{D_{\mathcal{X}\times \mathcal{L}}}$.

Based on the NSWs taxonomy developed by Trang et al., we construct the label set $L$, which includes the list of labels in Table \ref{tabII}. There are some differences in our NSWs taxonomy compared to the previous work. First of all, NTEL (telephone number, e.g., 0349663543) and NDIG (number as digits, e.g., 001097007189) are merged into a single tag NDIG because the tokens tagged with these tags have the similar verbalization so that the same converters can be used in the expansion. This change makes the model leaner and more efficient because it can eliminate the ambiguity in defining these two tags. Splitting into two separate tags is redundant.
In addition, some tags are removed, which are: ADD (address), CSEQ (read all characters), DURA (duration), NONE (ignored), and PUNC (punctuations).
The distinction between CSEQ and LSEQ (letter sequence) is unclear. Tokens corresponding to the CSEQ (e.g., ``xxx'') can adequately be captured under the LSEQ. The punctuation tokens respective to the PUNC (e.g., ellipses, quotation marks, dot, comma, etc.) are separated into individual tokens; they are treated as standard words. The tokens that expand to nothing respective to the NONE (e.g., ASCII art) are deleted in the pre-processing module. An address corresponding to the ADD tag is made up of many tokens, but most of them do not need to be expanded, e.g., ``\begin{otherlanguage}{vietnamese}
Số 144 Xuân Thuỷ, Cầu Giấy, Hà Nội\end{otherlanguage}'' (``144 Xuan Thuy, Cau Giay, Hanoi''). Therefore, we use other tags to label the tokens that need to be expanded, e.g., The NNUM tag for the ``144'' token. A further ablation is that of the DURA (duration). This tag defines a duration such as ``.'', ``-'' in telephone number (e.g., ``0966.3553.46'') or score (e.g., ``2-3''). 
In our tagging problem setup, the punctuations between these entities are not used as boundaries to split these entities into smaller tokens. Finally, four tags NQUA, NVER, ROMA, and MEA were added to capture the classes that we consider essential. The MEA (measurement, e.g., ``50kg'') and the NQUA (quarter, e.g., ``I/2022'') are added by the rapid proliferation of financial and economic articles on the internet. The NVER is used to distinguish version tokens from number tokens, e.g., ``4.0'' reads as ``\begin{otherlanguage}{vietnamese}bốn chấm không\end{otherlanguage}'' (``four dot zero'') instead of ``\begin{otherlanguage}{vietnamese}bốn\end{otherlanguage}'' (``four''). Likewise, the ROMA will instruct expanders to convert the roman numeral into its decimal verbalization. After reduction, our tag set used for the tagging models was 19 tags; detail can be found in Table \ref{tabII}.

To build a sequence tagging model, we conduct experiments with 3 methods: (1) A \textbf{CRF} model\cite{crf} with a hand-craft features dictionary that considers the characteristics of the tokens and surrounding tokens. The characteristics of each token include suffix and prefix, upper-cased or lower-cased, letter or digit number,
(2) \textbf{BiLSTM-CNN-CRF}\cite{bilstm-cnn-crf}: The CNN\cite{cnn} computes the character-level representation. The output representation vector of CNN is feed to the BiLSTM network\cite{lstm} to produce the final representation vector. The CRF layer takes the final vector and decodes the sequences of labels, 
(3) \textbf{BERT-BiGRU-CRF}\cite{bert-bigru-crf}: The presentation of input is obtained through a pre-trained EnViBERT model. Then the vector representation is given to the forward and backward Gated Recurrent Unit-BiGRU\cite{bigru} to capture the left and right context of the current word. The final representation vector is the concatenation of the left and right contextual vectors. A CRF built on top of BiGRU to predict sequences of labels.

\subsection{Non-Standard Word Normalization}

Rather than converting tokens straight from the written form into normalized spoken form, raw tokens will be used to determine to which label the token belongs. This predicted label will determine the type of normalization applied to the token. This allows us to partition the tokens and split up one massive conversion algorithm into 19 smaller ones - called the \textit{expanders}. The expanders define the kind of normalization applied to the token for a correct result. 
For instance, there are many ways to convert ``3-4'', the correct result might be ``\begin{otherlanguage}{vietnamese}
ba đến bốn\end{otherlanguage}'' (``three to four''). While, the class is NDAY label, the output should be ``\begin{otherlanguage}{vietnamese}ba tháng tư\end{otherlanguage}'' (``the third of April''). On other hand, the label is NSCR (a score in games), the correct output is ``\begin{otherlanguage}{vietnamese}ba bốn\end{otherlanguage}'' (``three four''). Each NSW follows transformation rules specific to the label the NSW belongs to. Before generating the normalized spoken form, the NSW is mapped into pattern classes, e.g., ``10/3/2000'' $\to$ ``\verb|dd/mm/yyyy|''. The NSW then will be tokenized into [``10'', ``/'', ``3'', ``/'', ``2000''], from that, the expander transforms each monotonic token from the unnormalized written into their corresponding spoken normalizations, e.g., [``\begin{otherlanguage}{vietnamese}mười\end{otherlanguage}'', ``\begin{otherlanguage}{vietnamese}tháng\end{otherlanguage}'', ``ba'', ``năm'' ``\begin{otherlanguage}{vietnamese}hai nghìn\end{otherlanguage}'']. 
This process applied to NSWs in the \textbf{\textit{Number group}} and \textbf{\textit{Other group}} (except URLE class). For abbreviation tokens (LABB) and foreign word tokens (LWRD) in the \textbf{\textit{Letter group}}, we look them up in two pre-built dictionaries and get the verbal form. The remaining letter sequence tokens (LSEQ) are generally converted by padding the alphabetical characters with spaces and keeps the character verbatim form e.g., ``VTV'' $\to$ ``V T V''. 

Lastly, we handle the problem of URLE separation. URLE class consists mainly of emails, URLs, hashtags, and contact names, e.g., ``\#anhkhanh'', ``phongdaotao@vnu.edu.vn''. Our approach first split the prefixes (e.g.,``\#'') and the suffix (e.g., ``@vnu.edu.vn'') of the email addresses, hashtags and URLs into individual tokens and mapping into the dictionary to expand. The primary tokens are multi-syllabic tokens or concatenations of syllables without space separate e.g. ``anhkhanh'', ``phongdaotao''. To break down these main tokens, we propose to use a forward lexicon-based maximum matching algorithm (FLMM). From the start index of the word, FLMM employs a sliding window size $k$ (with $k$ decreasing) to slide through the word from left to right and look up this $k$-size token in the lexicon. Suppose the token is found in the lexicon. In that case, the algorithm marks a boundary at the end of the longest token and segments that sequence of this word, then begins the same longest search starting at the character following the boundary. If the token is not found in the lexicon, FLMM increases the start index of the sliding window and starts searching at the next character of the word.
The algorithm is described in more detail in Algorithm 1.
\begin{algorithm}
 \caption{Forward Lexicon-based Maximum Matching}
 \begin{algorithmic}[1]
 \renewcommand{\algorithmicrequire}{\textbf{Input:}}
 \renewcommand{\algorithmicensure}{\textbf{Output:}}
 \REQUIRE A lexicon set $L$ and a string name $s$
 \ENSURE  A separate-syllable string name
  \STATE result $\leftarrow$ []\\
  \STATE tokens $\leftarrow$ \textit{list}($s$)\\
  \STATE minWindow $\leftarrow$ 1\\
  \STATE maxWindow $\leftarrow$ \textit{len}(tokens)\\
  \STATE startIdx $\leftarrow$ 0\\
  \STATE gotcha $\leftarrow$ \textbf{False}
  \WHILE {startIdx $\leq$ \textit{len}(tokens) - minWindow}
  \STATE gotcha = \textbf{False}
    \FOR{wSize \textbf{in} \textit{range}(maxWindow, minWindow-1, -1)}
        \STATE endIdx = startIdx + wSize - 1
        \STATE candidate = tokens[startIdx:endIdx+1]
        \STATE candidateString = ``".\textit{join}(candidate)
        \IF{candidateString \textbf{in} $L$}
            \STATE result.\textit{append}(`` " + candidateString + `` ")
            \STATE startIdx = endIdx + 1
            \STATE gotcha = \textbf{True}
            \STATE \textbf{break}
        \ENDIF
    \ENDFOR
    \IF{\textbf{not} gotcha}
        \STATE result.\textit{append}(tokens[startIdx])
        \STATE startIdx = startIdx + 1
    \ENDIF
  \ENDWHILE
  \IF{startIdx $<$ \textit{len}(tokens)}
    \STATE result.\textit{append}(tokens[startIdx])
  \ENDIF
 \RETURN `` ''.\textit{join}(result)
 \end{algorithmic} 
 \end{algorithm}

\section{Evaluation}

\subsection{Data Preparation and Statistics}

\begin{figure}[h!]
  \includegraphics[width=0.45\textwidth]{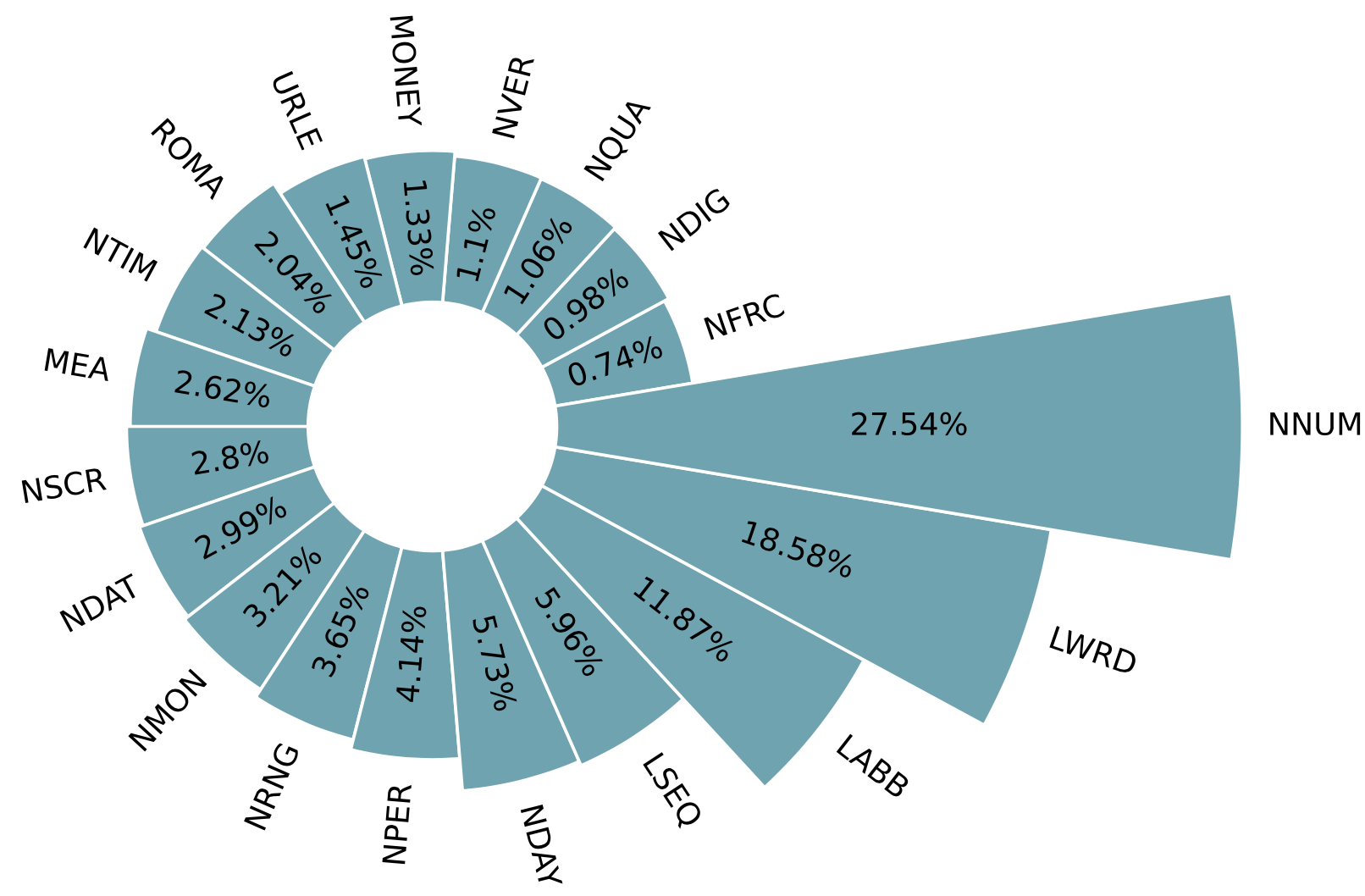}
  \caption{Label distributions}
  \label{fig2}
\end{figure}

To build dataset on NSWs and conduct experiments, documents crawled from the vnexpress.net  website with many topics (including news, business, finance, sport, etc.). Since, we select and annotated sentences containing NSW. We obtained a dataset consisting of 5819 labeled sentences (194868 tokens), which contained 13472 NSWs, with an average of 2.31 NSWs per sentence. The statistics for data are shown in detail in Table \ref{tabIII}, and the distribution of NSW labels is detailed in Figure \ref{fig2}. We randomly divided the dataset into a training set and a testing set with the number of examples 3991 and 1828 respectively. In addition, to perform experiments to evaluate the effectiveness of the FLMM algorithm for URLE separation, we collected 9796 contact names from phone contact names. Besides, a lexicon is collected from 3 sources, Vietnamese name databases\footnote[1]{https://github.com/duyet/vietnamese-namedb}, Vietnamese wordlists\footnote[2]{https://github.com/duyet/vietnamese-wordlist}, and Underthesea dictionaries\footnote[3]{https://github.com/undertheseanlp/underthesea}. 
$BIO$ schema is used to tag the phrase NSW. $B$--prefix indicates that the token is the beginning of a NSW, and $I$--prefix indicates that the token is inside a NSW. $O$ tag indicates that the token belongs to a standard word. The output layer of the tagging model produces 39 labels ($O$, $B$--$X$, $I$--$X$, where $X$ is in a set of 19 tags).


\begin{table}[htbp]
\caption{Data Statistics}
\begin{center}

\begin{tabular}{|l| c|} 
 \hline
 \textbf{Statistics} & \textbf{Values}\\
 \hline
  \#Sentences & 5819\\
  \#Tokens & 194868\\
 \#Token/sentence & 33.48\\
 \#NSWs & 13472\\
 \#NSW/sentence & 2.31\\
 \hline
\end{tabular}\\[0.2cm]

\resizebox{0.49\textwidth}{!}{
\begin{tabular}{|l| c|c|c|} 
 \hline
 \textbf{Group} & \textbf{Distribution for data} & \textbf{Distribution for training} & \textbf{Distribution for testing}\\
 \hline
  \textbf{\textit{Number}} & 56.50\% & 56.32\% &  56.9\%\\
  \hline
   \textbf{\textit{Letter}} & 36.42\% &36.72\% &35.74\%\\
  \hline
   \textbf{\textit{Other}}&7.08\%  &6.96\% &7.36\%\\
   \hline
   \textbf{Total}&100.00\%  &100.00\% &100.00\%\\
 \hline
\end{tabular}
}

\label{tabIII}
\end{center}
\end{table}

\subsection{Model Setup}

Conditional Random Field is training with gradient descent with L-BFGS method, the coefficient for $L_1$ and $L_2$ regularization is set 0.1, The maximum number of iterations for optimization algorithms is 100. With the BiLSTM-CNN-CRF model, we use 300-dim word2vec pre-trained embedding for the Vietnamese language from \cite{word2vec} for word embedding. The other hyperparameters are setup as in \cite{bilstm-cnn-crf}. Model BERT-BiGRU-CRF with EnViBERT for word representation has 4 BiGRU layers and the hidden state of BiGRU is 512. Model optimization using Adam\cite{adam}, the batch size is 64, and training with 30 epochs.

\subsection{Evaluation Metric}

We use \textit{Precision ($P$)}, \textit{Recall ($R$)}, \textit{$F_1$} scores for NSW detection evaluation and \textit{Sentence Error Rate (SER)} for NSW normalization evaluation. \textit{SER} is the number of incorrect normalized sentences divided by the total number of sentences.

\subsection{Results and Analysis}

\begin{table}[htbp]
\caption{Precision, Recall, and $F_1$ scores in CRF, BiLSTM-CNN-CRF and BERT-BiGRU-CRF}
\begin{center}
\resizebox{0.49\textwidth}{!}{
\begin{tabular}{|l|c|c|c|c|c|c|c|c|c|}
\hline
\multicolumn{1}{|c|}{} & \multicolumn{3}{c|}{\textbf{CRF}} & \multicolumn{3}{c|}{\textbf{BiLSTM-CNN-CRF}}&\multicolumn{3}{c|}{\textbf{BERT-BiGRU-CRF}}\\
\cline{2-10}
\multicolumn{1}{|l|}{\textbf{Label}}& \textit{P} & \textit{R} & $F_1$ & \textit{P} & \textit{R} & $F_1$ & \textit{P} & \textit{R} & $F_1$\\
\hline
LABB&0.91&      0.94 &     0.93&\textbf{0.97}&      0.92&      0.94&0.95&\textbf{0.97}&\textbf{0.96}\\
LSEQ&\textbf{0.93} &     0.89  &    \textbf{0.91}&0.76 &     0.81 &     0.78&0.84&\textbf{0.91}&0.87   \\
LWRD &\textbf{0.90} &     0.76  &    0.82& 0.83&      0.87&      0.85 &0.89&\textbf{0.90}&\textbf{0.90}\\
MEA &\textbf{0.99}   &   0.72    &  0.84& 0.83  &    0.80  &    0.81 &0.93&\textbf{0.95}&\textbf{0.94}\\
MONEY&\textbf{1.00}   &   0.90    &\textbf{0.95}& 0.89    &  0.89    &  0.89 &0.92&\textbf{0.90}&0.91\\
NDAT& 0.97    &  0.86     & 0.91& 0.95   &   0.99   &   0.97 &\textbf{0.98}&\textbf{1.00}&\textbf{0.99}  \\
NDAY& 0.93     & 0.91      &0.92& 0.97    &  0.95    &  0.96 &\textbf{0.97}&\textbf{0.99}&\textbf{0.98}\\
NDIG& \textbf{0.94}      &0.36&      0.52& 0.76     & 0.66     & 0.71 &0.74& \textbf{0.80}&      \textbf{0.77} \\
NFRC& 0.88&      0.68 &     0.76& 0.85      &0.71      &0.77 & \textbf{0.92}&      \textbf{0.71}&      \textbf{0.80}\\
NMON& 0.98 &     0.99  &    0.98& \textbf{0.99}&      0.99&      \textbf{0.99} &0.98 &     \textbf{0.99}  &0.99\\
NNUM& 0.94  &    0.95   &   0.95& \textbf{0.98} &     0.95 &     0.96 &0.97 &     \textbf{0.99}  &\textbf{0.98}\\
NPER& \textbf{0.99}   &   0.85    &0.92& 0.96    &  0.96    &  0.96 &0.95 &     \textbf{1.00}      &\textbf{0.98}\\
NQUA& \textbf{1.00}    &  \textbf{1.00}     & \textbf{1.00}& 1.00   &   0.93   &   0.97 &1.00 &     0.96      &0.98\\
NRNG& 0.90     & 0.78      &0.83& 0.86    &  0.92    &  0.89 &\textbf{0.91} &     \textbf{0.95}     & \textbf{0.93}\\
NSCR& 0.97      &0.91&      0.94& 0.96     & 0.97     & 0.97 &\textbf{0.97}  &    \textbf{0.98}    &  \textbf{0.98}\\
NTIM& 1.00&      0.92 &     0.96& 1.00      &0.94      &0.97 &\textbf{1.00}   &   \textbf{1.00}   &   \textbf{1.00}\\
NVER& \textbf{0.98} &     \textbf{0.94}  &    \textbf{0.96}& 0.98&      0.92&      0.95 &0.92&      0.73  &    0.81\\
ROMA& \textbf{0.97}  &    \textbf{1.00}   &   \textbf{0.99}& 0.97 &     0.99 &     0.98 &0.96 &     0.99 &     0.97\\
URLE& \textbf{1.00}   &   \textbf{0.91}    &  \textbf{0.95}& 0.95  &    0.84  &    0.89 &0.92  &    0.87&      0.89\\
\hline\hline
$Average_{micro}$& 0.94     & 0.88 &     0.91& 0.92    &  0.92    &  0.92 &\textbf{0.94}  &    \textbf{0.95}     & \textbf{0.95}\\
$Average_{macro}$& 0.91  &    0.81    &  0.85& \textbf{0.92}     & \textbf{0.89} &     \textbf{0.91} &0.89  &    0.88     & 0.88\\
\hline
\end{tabular}}
\label{tabIV}
\end{center}
\end{table}
\textit{Tagging model performances}: 
Table \ref{tabIV} shows the performance of three tagging models. In terms of $P$, $R$, and $F_1$, BiLSTM-CNN-CRF model achieved $P$ = 92\%, $R$ = 92\% and $F_1$ = 92\% and has 4\% and 1\% higher, and 3\% lower to compare with baseline CRF model respectively. The tagging model with BERT for input representation gives the best results with $P$ = 94\%, $R$ = 95\%, and $F_1$ = 95\%. Compared with BiLSTM-CNN-CRF model, the increase in $P$, $R$, and $F_1$ scores is 2\%, 3\%, and 3\%, respectively. Compared with CRF model, the increase is 7\% and 4\% in terms of $R$ and $F_1$, respectively, and achieve a similar $P$ = 94\%. Two neural-based models show their powerful semantic representation, which allows the word vectors to incorporate context information better. The CRF model with rich feature contexts also shows intense competition with neural-based models due to the limited labeled training data. 

\begin{table}[htbp]
\caption{normalization results}
\begin{center}
\begin{tabular}{|r|c|} 
 \hline
 \textbf{Experiments}  & \textbf{SER}\\
 \hline
  CRF + Normalization &   8.15\% (149/1828)\\
  \hline
   BiLSTM-CNN-CRF + Normalization &   7.11\% (130/1828)\\
  \hline
   BERT-BiGRU-CRF + Normalization &   6.67\% (122/1828)\\
 \hline\hline 
 URLE separation &   9.16\% (897/9796)\\
 \hline
\end{tabular}
\label{tab4.5}
\end{center}
\end{table}

\textit{Normalization results}: The result of the NSW normalization phase is shown in Table \ref{tab4.5}. With the CRF tagger, the number of incorrect sentences was 149 out of 1828 sentences, corresponding to a \textit{SER}=\textbf{8.15\%}. The number of incorrect sentences of the NSW normalization with config BiLSTM-CNN-CRF and BERT-BiGRU-CRF taggers are 130/1828 and 122/1828, corresponding \textit{SER}=\textbf{7.11\%} and \textit{SER}=\textbf{6.67\%}, respectively. 

The rule-based normalizer can be quickly implemented and has a low inference time. That makes it well suited for use in real-time applications. However, the drawback of the TTS normalization system with a rule-based normalizer is that it is poor in generalization and requires more effort to maintain if something goes wrong. Another mistake is that the tagger assigns tokens to the wrong class, causing the expansion to go wrong. There are a few cases where the tagging model identifies the LWRD token as a LABB tag, but it still gives the correct result when expanded.
For example, ``NATO'' is defined as a foreign word (LWRD), but the model's tagging is LABB; the pre-built LABB expansion dictionary has ``NATO'' and then expands to ``na tô", so the expansion is still correct.
In addition, there are other NSW expanders errors, such as some abbreviations that are not expanded correctly because of ambiguity. Otherwise, most errors are due to NDIG tokens (telephone/digits number) being detected with the incorrect label as NNUM tokens (numbers), leading to incorrect expansions for NDIG tokens. The reason can be explained by the relatively small number of samples containing NDIG, and the patterns of these tokens are similar to the NNUM token.

\begin{table}[htbp]
\caption{Some incorrect predictions of FLMM}
\begin{center}
\resizebox{0.49\textwidth}{!}{
\begin{tabular}{|l|l|l|} 
 \hline
 \textbf{Contact name}  &\textbf{Prediction}  & \textbf{Ground truth}\\
 \hline
   Vinasun & Vin a sun & Vi na sun\\
   \hline
  bachoa& bach oa& bac hoa\\
  \hline
   Vinasun & Vin a sun & Vi na sun\\
 \hline
 chacathu@nhatrang & cha cat hu @ nhat rang & cha ca thu @ nha trang\\
  \hline
  Đạt@vietnamwork & Đạt @ Viet nam w o r k &	Đạt @ Viet nam work \\
  \hline
\end{tabular}}
\label{tab4.6}
\end{center}
\end{table}

\textit{URLE separation}: The error rate of the URLE separation is \textit{SER}=\textbf{9.16\%}. FLMM has wrongly split 897 over 9796 contact names. We have observed two main errors. First, the ambiguity problem is caused by the greedy nature of the FLMM algorithm when forward sliding from left to right is used. For example, ``vinasun'' must be separated into ``vi na sun'' but the lexicon has ``vin'', so FLMM split it into ``vin a sun''. Second, the lexicon does not cover English tokens, so it is split into individual characters, e.g., ``work'' $\to$ ``w o r k''.
\section{Conclusions}

In summary, we proposed a new two-phase approach for the Vietnamese TTS normalization problem, which combines the model-based NSW tagger and the rule-based NSW normalizer. The experiment results of tagging phase archived high accuracy show a practical approach to handling the semantic ambiguity of NSW. In the normalization phase, we proposed a solution to split URL, email, hashtag, and contact name into separate pronounceable syllable tokens using a forward lexicon-based maximum matching algorithm. Our approach is a promising study that can be applied to the real-world TTS.

\end{document}